\documentclass[12pt]{article}

\usepackage{arxiv}
\usepackage[utf8]{inputenc} 
\usepackage[T1]{fontenc}    
\usepackage{hyperref}       
\usepackage{url}            
\usepackage{booktabs}       
\usepackage{amsfonts}       
\usepackage{nicefrac}       
\usepackage{microtype}      
\usepackage{lipsum}
\usepackage{epsfig}  
\usepackage{amsmath}
\usepackage{algorithm}
\usepackage{algorithmic}   
\usepackage{graphicx}
\usepackage{color}
\usepackage[dvipsnames]{xcolor}
\usepackage{amssymb}  
\usepackage[algo2e]{algorithm2e}
\usepackage{tikz}
\usepackage{cite}
\usepackage{array}
\usepackage{romannum}
\usepackage{xcolor} 
\usepackage{soul}
\usetikzlibrary{matrix,chains,positioning,decorations.pathreplacing,arrows}

\title{
Cell A* for Navigation of Unmanned Aerial Vehicles in Partially-known Environments
}

\author{
  Wenjian Hao\thanks{Graduate Researcher, \tt\small whao@g.clemson.edu}, \
  Rongyao Wang\thanks{Graduate Researcher, \tt\small rongyaw@g.clemson.edu}, \
  Alexander Krolicki\thanks{Undergraduate Researcher, \tt\small akrolic@g.clemson.edu},\
  Yiqiang Han\thanks{Research Assistant Professor, \tt\small yiqianh@g.clemson.edu} \thanks{Corresponding Author}\\
  \textit{Department of Mechanical Engineering}\\
  \textit{Clemson University}\\
  Clemson, SC 29634 USA \\}

\begin{document}
\maketitle

\begin{abstract}
Proper path planning is the first step of robust and efficient autonomous navigation for mobile robots. Meanwhile, it is still challenging for robots to work in a complex environment without complete prior information. This paper presents an extension to the A* search algorithm and its variants to make the path planning stable with less computational burden while handling long-distance tasks. The implemented algorithm is capable of online searching for a collision-free and smooth path when heading to the defined goal position. This paper deploys the algorithm on the autonomous drone platform and implements it on a remote control car for algorithm efficiency validation. 
\end{abstract}

\bigskip
\section{Introduction}

Unmanned Aerial Vehicle (UAV) is a critical component in the ecosystem of autonomous systems designs. UAVs have unparalleled flexibility in the three-dimensional space, creating novel applications that are impossible for autonomous ground vehicles (AGV) to accomplish. These extra degrees of freedom are coupled with added complexity to the dynamics of the system. UAVs navigate primarily in unstructured environments but in some cases they must navigate structured environments where a clear path is not always present. As a vital step for autonomous navigation, path planning must schedule a smooth path and avoid obstacles in real-time for UAVs and AGVs alike. In order to solve the UAVs path planning problem, various algorithms have been proposed to obtain an energy-efficient and obstacle-free route. The existing algorithms can be classified into the following categories: sample-based algorithms, heuristic-based algorithms, and intelligent bionic algorithms. For more algorithms, details refer to \cite{class, mldrone, reviewdrone}.

This study focuses on the application of an extended A* algorithm which is also a heuristic-based algorithm \cite{Ferguson-2005-9201}. Since the sampling-based approach cannot guarantee an optimal solution, we do not want to put our mission-critical path planning tasks up to chance or luck of the draw. The heuristic-based algorithms have been widely implemented in various fields such as autonomous driving \cite{adriving}, autonomous parking \cite{parking}, etc. The standard A* algorithm's basic idea depends on a global or local heuristics table to track the best candidate node with the lowest computed heuristic cost. The algorithm works efficiently in a reasonably scaled map with known obstacles. To achieve dynamic re-planning in the vicinity of dynamic obstacles in partially known environments, several A* variants have been employed such as D*, Anytime A* (ARA) \cite{dstar, ara}, etc.

Although A* and its variants have been successfully applied to various mobile robots in two-dimensional structured environments, practical use in UAVs path planning still suffers from the increased degrees of freedom as well as a lack of explicit prior knowledge of the environment. For example, micro-scale/small-scale UAVs in urban environments bring about challenges associated with complexity from the highly structured environment, noisy signals due to EMI, and a need for energy efficiency due to battery constraints for flights covering long distances of unexplored areas. In many practical use cases of UAVs, GPS signals are limited, especially for indoor or rural areas. Therefore, we need to mainly rely on data collected from distance sensors to obtain a local map, rather than an accurate plan available for AGVs on roads with high-definition maps. Besides, 3-D environments are often too large to achieve accurate global planning and inevitable bias due to position measurement error, execution bias, etc. and adds complexity.

To overcome the aforementioned problems, this study proposes two extensions to the A* and its variants deployed in three-dimensional space. First, in real-time, we prioritize a local cubic search cell as the drone's sampling space and pick nodes with the lowest cost at each layer to get a 'cell path.' In this way, we can re-plan a cell path at each time step to reduce the execution bias without accumulating error. Secondly, to avoid aggressive path smoothing, we introduce an extra cost for the distance between the current position and a 3-D straight line connecting the starting point and goal point.  In this way, we can drive the drone close to the theoretic shortest path and make long-distance path planning in an energy- and time-efficient manner.

\begin{figure}[h!t]
    \centering
    \includegraphics[width=0.72\textwidth]{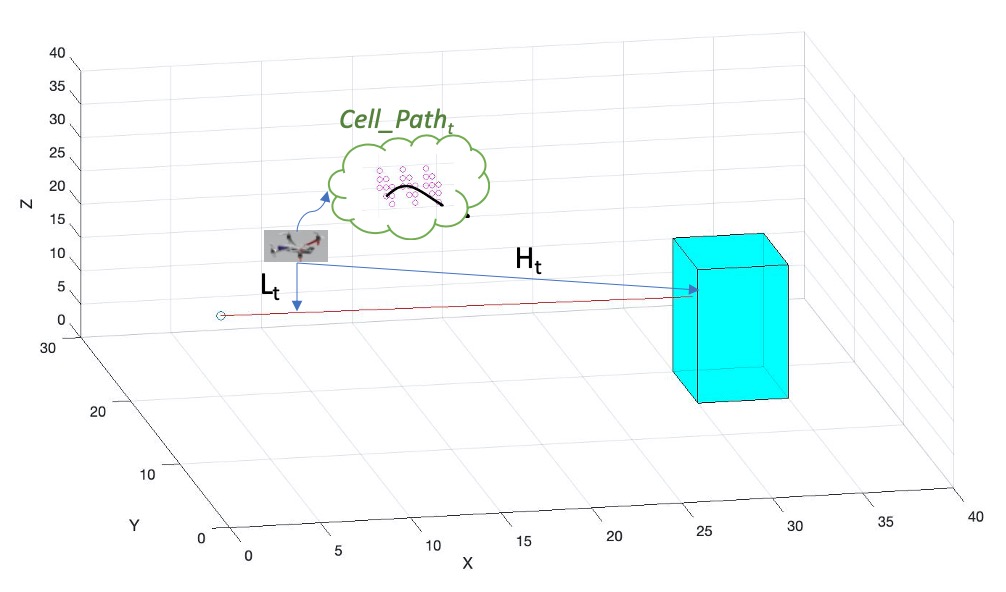}
    \caption{Path planning in 2-D and 3-D scenarios}
    \label{fig:scheme}
\end{figure}

The experimentation of this algorithm is divided into two parts. The first part is conducted within a customized 2-D simulation environment. For this simple 2-D example, the dynamical system model is defined as a bicycle model for an AGV, which is essentially a simplified version of the Ackermann steering model for low-speed application \cite{pure_pursuit}. The 2-D simulation was implemented in MATLAB\cite{MATLAB:2019b} for demonstration purposes. In this case, 2D testing is implemented based on our modification of hybrid A* for a non-holonomic vehicle. The second example is based on a more complex UAV model simulated in a 3-D GAZEBO virtual environment \cite{gazebo}. It utilizes a simultaneous localization and mapping (SLAM) algorithm for UAVs localization \cite{slamm} and a stereo camera to create a point cloud map of unknown obstacles. A schematic of the environment is shown in Figure \ref{fig:scheme}.

This paper is organized as follows: Section \Romannum{2} briefly introduces the A* algorithm and its major variants and presents details of the extensions we propose. Section \Romannum{3} shows the algorithm verification in the 2-D world by simulating the algorithm on an RC car with constraints in a 2-D Matlab simulator. Section \Romannum{4} exhibits the experiment execution on the drone platform in both simulator and the real world with the results. Finally, section \Romannum{5} concludes the paper.

\bigskip
\section{Extension of Hybrid A* Algorithm}
In this section, we briefly present the A* algorithm with its variants, and our extended version of the A* algorithm: Cell A*. We will also discuss the different treatments when applying the proposed algorithm to AGV with a bicycle model on a 2-D plane and a UAV with holonomic behavior in 3-D space.


\subsection{Hybrid A*}

\begin{figure}[h!t]
    \centering
    \includegraphics[width=0.6\textwidth]{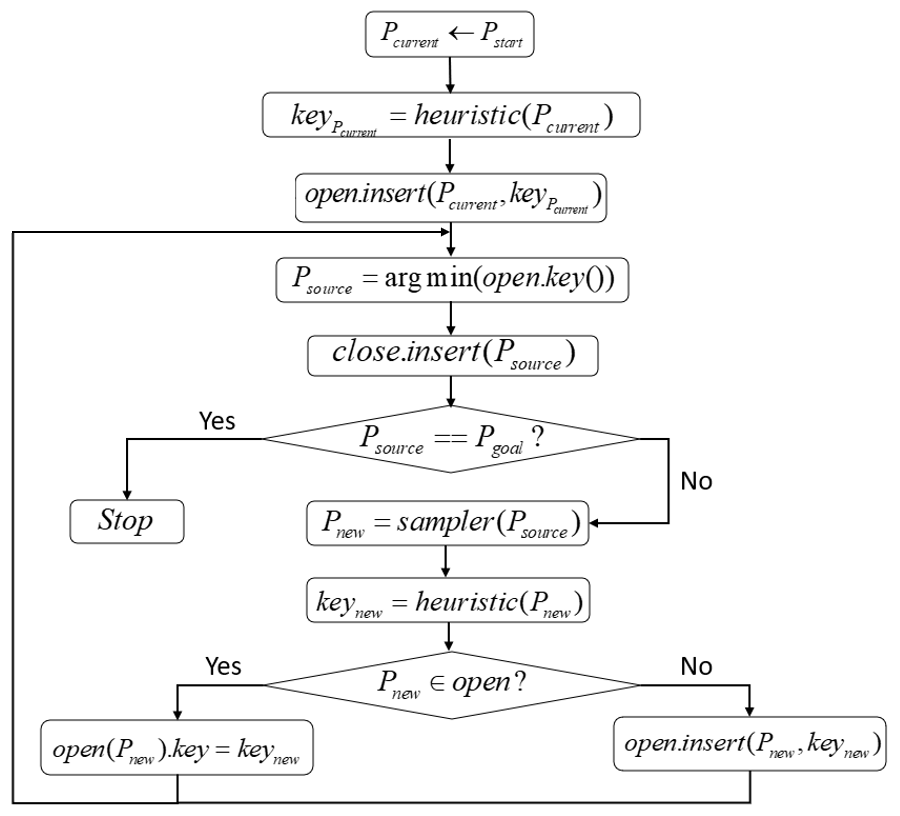}
    \caption{Schematics of hybrid A* in steps}
    \label{fig:hybrid_A_star_demo}
\end{figure}

In real-world applications, standard A* has a disadvantage that the planned path could be infeasible for robots with differing kinematic constraints. In practical application, Hybrid A* is proposed to include non-holonomic constraints during expanding path search, as shown in unstructured outdoor environments \cite{adriving} and unknown semi-structured environments \cite{hybrid_a_stanford}. A standard hybrid A* flow chart is shown in Figure \ref{fig:hybrid_A_star_demo}.

In Figure \ref{fig:hybrid_A_star_demo} the heuristic function $H$ in Equation \ref{eqn.standard_cost} maintains a tracking record of two distances. The first part is $f_t$, the euclidean distance from the current state to the goal state. $g_t$ is the euclidean distance from the current state to the start state. $w_t$ is the hyperparameter with a value between 0 and 1, which balances the priority between chasing the final goal and finding an optimal path.
\begin{equation}\label{eqn.standard_cost}
    \begin{gathered}
        f_t = ||P_{t}(x_t, y_t, z_t) - P_{goal}(x_g, y_g, z_g) ||_2\\
        g_t = ||P_{t}(x_t, y_t, z_t) - P_0(x_0, y_0, z_0) ||_2\\
        H_t = f_t + w_g \cdot g_t
    \end{gathered}
\end{equation}

Hybrid A* generates a path based on the global map as well as a predefined non-holonomic sampler with a sampling direction aligned with the goal position. Before implementing hybrid A* planning, all environment information is assumed to be available to maintain a rapid search on a small scale map. Yet in the real-word application, planning across the entire map requires a large amount of memory storage and high latency cost. Furthermore, the environment information may not be fully known. Hence, for practical use purposes, when the robot is in a semi-known environment, it is necessary to navigate and re-plan its path by constructing a map and generating a path simultaneously \cite{hybrid_a_stanford}, which can, in turn, lead to tens of thousands of nodes of information.

With sufficient computing power, the Hybrid A* is guaranteed to achieve optimal path planning for a given problem. However, in applications for micro/small scale autonomous vehicles, the computing power is limited, and real-time online searching capability is more of a priority. Therefore, we seek an extension to the current algorithm for an efficient algorithm suitable for edge computing on a mobile robot.

\subsection{Extension of A* algorithm}
To maintain hybrid A* exploratory capability while shrink the minimum necessary size of the map and number of nodes stored, this paper proposes an extension of the hybrid A* algorithm that can dissect the global hybrid A* into local hybrid A* within a cell that has an adaptive size and ensures the smoothness of the path. 

The proposed algorithm shares the same backbone of Hybrid A* while expanding the adaptive searching capability into the 3-D neighboring nodes. The details of the algorithm consist of the following steps:
Considering the focus of this paper is the drones navigation, the algorithm should sample its neighbor nodes in all the directions. 

\begin{enumerate}
    \item  Generate Search Nodes:\\ For predictive nodes sampling, we choose a cubic cell for the drone: $Cell_{t}(cell size, grid size)$, where the $cell size$ is the number of nodes on each cell side, the $grid size$ is the length of each grid in the cell. In this way, the cell can cover all the directions and expand the range of searching if necessary. The search cell ($Cell^{search}_t$) of position $P_t$ at time $t$ would be $Cell_{t}(cell size, grid size)+P_t$
    \item Evaluate the costs of all the valid nodes in $Cell^{search}_t$:\\ To reduce the computation of the evaluation, we compare the adjacency cells with the real-time updated local occupancy map. We only take the valid cell points, which is not in the obstacles point cloud set. The cost contains two parts. One is an A*-like heuristic cost $H_t$, which tracks the distance between the current position ($P_t$) and goal position ($P_{goal}$); the other cost $L_t$ is the distance between the current position ($P_{goal}$) and the Euclidean shortest path which connects the initial position and the goal position. While $H_t$ pushes the mobile robots close to the goal position, $L_t$ ensures a theoretic shortest path of the robots and compensates the bias generated by the path following, especially for drones in complex outdoor environments. The total cost ($J$) is shown in Equation \ref{eqn.cost}, $w1$ and $w2$ is weights of $H_t$ and $L_t$.
    \begin{equation}\label{eqn.cost}
    \begin{gathered}
        J = w_1 H_t + w_2 L_t
    \end{gathered}
    \end{equation}
    Where $H_t$ is in Equation \ref{eqn.hc}
    \begin{equation}\label{eqn.hc}
    \begin{gathered}
        H_t = ||P_{t}(x_t, y_t, z_t) - P_{goal}(x_g, y_g, z_g) ||_2
    \end{gathered}
    \end{equation}
    To track the cost term $L_t$, we require a partially known map which enables us initialize the starting point $P_0(x_0, y_0, z_0)$, goal position $P_{goal}(x_g, y_g, z_g)$ and obtain current position $P_{t}(x_t, y_t, z_t)$ with Simultaneous Localization and Mapping (SLAM) or GPS. The cost term $L_t$ is defined in Equation \ref{eqn.lc}.
    \begin{equation}\label{eqn.lc}
    \begin{gathered}
        3D\ line: \frac{x - x_0}{x_g - x_0} = \frac{y - y_0}{y_g - y_0} = \frac{z - z_0}{z_g - z_0}\\
        \\
        L_t = |distance(P_t, line)|
    \end{gathered}
    \end{equation}
    
    \item Verify Obstacles: 
        \begin{enumerate}
        \item If front obstacles on path are out of avoidance range: Set $Cell^{search}_t(3, bigstep)$, where $cellsize = 3$, $gridsize = big step$ means it is a one layer cell with 3 nodes on each edge so as to enable a fast path planning. Meanwhile,
        $J = w_1 H_t + w_2 L_t$
        \item Else, if it is able to pass the obstacle: $J = w_1 H_t + w_2 L_t$\\ if not: $J = w_1 H_t - w_2 L_t$
        \item We allow the sign of $L_t$ term to change to balance the need to find the shortest path based on global heuristics and encourage local replanning to avoid obstacles along the shortest path. To avoid unforeseen obstacles or dynamic obstacles along the global path, the robot may need to deviate from the shortest path and pivot to exploration in neighboring cells. We allow the sign of $L_t$ in heuristic function to reverse to encourage robot navigating away from the global planned path. 
        \end{enumerate}
    
    \item {Search an optimal path in the search cell $Cell^{search}_t$: \\Set $cellpath$ list, in each layer of the search cell, select the node with the lowest cost ($N_t = argmin_{node}J(Cell^{search}_t$), then pass the lowest cost node to the cell path set ($cellpath \leftarrow N_t$), finally choose the first $cellsize - 1$ nodes as the cell path for safety sake at time step $t$.
    The planned path ($path_t$) at time step $t$ would be: $path_t \leftarrow cellpath_{1...(cellsize-1)}$.}
    
    \item Check for Constraints:\\
    Integrate constraints into the cost function for some mobile robots with kinematic limitations like RC cars steering. 
    \item Check for Arrival: If the distance between the current position and goal position is acceptable, STOP.
\end{enumerate}
The summary of the algorithm is shown below:
\DontPrintSemicolon
\begin{algorithm}[h!t]
\caption{Cell A* for UAVs in 3D environment}
\SetAlgoNoLine
\KwIn{Current state $P_t(x_t, y_t, z_t)$ $\leftarrow$ SLAM algorithm\\ \ \ \ \ Obstacles set  $\leftarrow$ points cloud from stereo camera}
\KwOut{Planned trajectory in the search cell at each time step $t$}

\textbf{Initialization}

    {\begin{enumerate}
        \item Set start state $P_0(x_0, y_0, z_0)$ and goal state $P_{goal}(x_g, y_g, z_g)$
        \item Set starting node = $node_0$
        \item Heuristic function H($P_t, P_{goal}$), L($P_{goal}, P_0$)
        \item Define nodal cost function $J(Cell_t^{search}, Obstacles)$\\
        \For{node in $Cell_t^{search}$}{
        
            \Indp SafeToPass = node $\notin$ $Obstacles\ set$? 1:-1\\
            \Indp $J_{node} = w_1 H_t + w_2 L_t\times SafeToPass$
        }
        \Indm
    \end{enumerate}}

\While{goal reached == false}
{   
    \eIf {obstacles on path $\notin$ search range} 
    {
        $Cell_t^{search}$ = $genSearchCell_{node}(3,gridsize)$ + $P_t$\\
        $Cell\ Path \leftarrow node = argmin_{node}J(Cell_t^{search}, Obstacles)$
    }
    {
        \For{i in range ($cell size$)}
        {
        $Cell_t^{search} = genSearchCell_{node}(i,gridsize) + P_t$ \\
    $Cell\ Path \leftarrow node = argmin_{node}J(Cell_t^{search}, Obstacles)$}
    }
}
\Return {Cell Path}

\end{algorithm}

\bigskip
\section{Algorithm Planning Efficiency Validation in 2-D Environment}
In this section, we validate the Cell A* in a MATLAB 2-D simulation environment \cite{MATLAB:2019b}. The entire validation consists of three parts. Firstly we implement it with non-holonomic constraints to mimic the real car driving scenario. We then compare the Cell A* with constraints with the local hybrid A*, finally, we compare the proposed Cell A* algorithm with the D* lite algorithm, which can ensure the shortest planned path in terms of the planned path length and planning time. To get a fair benchmark, we do not optimize any of these comparison algorithms.

\subsection{Cell A* Validation with kinematic constraints}
\begin{figure}[h!t]
\centering
\includegraphics[width=0.36\textwidth]{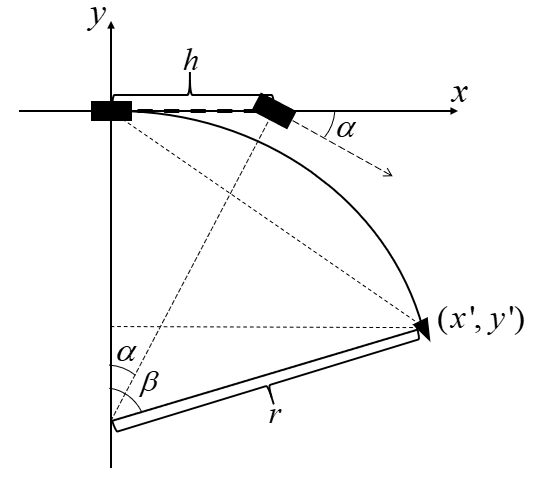}
\includegraphics[width=0.4\textwidth]{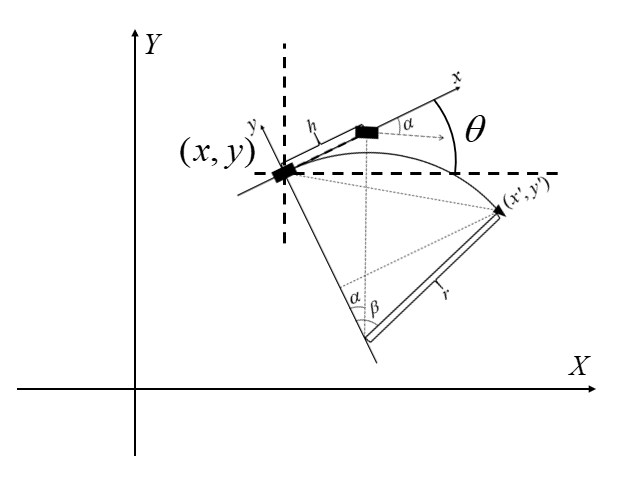}
\caption{Ackermann Sampler Demonstration}
\label{fig:ackermann}
\end{figure}

Unlike a drone's motion being holonomic, a car-like robot's motion is restricted by Ackermann steering kinematic model. Considering a minimum functional model \cite{pure_pursuit}, the vehicle states can be represented by $(x,y,\theta,d)$, where $x, y$ is coordinate position on a 2D plane, $\theta$ is the current yaw angle of the car with respect to x coordinate axis, and $d$ represents the direction of robot's motion. For simplicity, The rear axle is considered as the center of rotation. A schematic of the model setup is shown in Figure \ref{fig:ackermann}.

The expansion of the search cell is governed by two parameters, $l$, and $d$. $l$ control the distance range of the sampling cell edge in the 2D plane. A larger sampling distance $L$ means sampler can reach further on a 2D plane, thus increase exploratory performance. Change of $d$ can change the growth direction of sampling cells. Similar to the purpose of reversing the sign of $L_t$ in nodal cost function, changing the sign of $d$ also serves the purpose of encouraging the exploration behavior of the agent when an obstacle is detected along the global planned path. During the simulation, to further promote exploration behavior, the sampling steps within one search are also increased if local/dynamic obstacles are detected.

To accommodate the need for local replanning, which comes with the constrain of nonholonomic behavior, $g_t$ in the nodal cost function $J$ no longer tracks distance to the global start state, but rather the distance to the start state of the current sampling start position. In this case, the global start position got abandoned after first sampling, hence reducing the possibility of circling back to the original state when encountering obstacles.

\begin{figure}[h!t]
    \centering
    \includegraphics[width=0.46\textwidth]{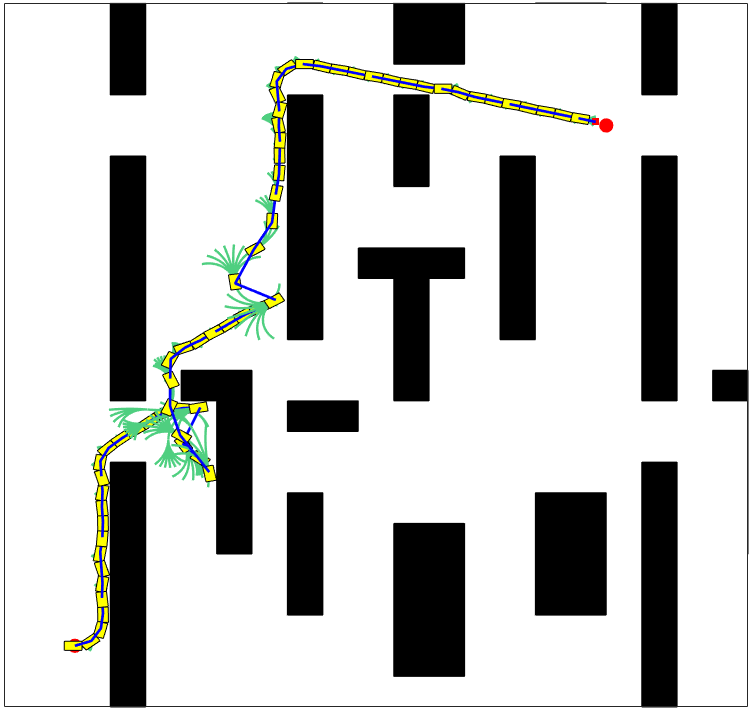}
    \caption{Cell A* real-time planning process}
    \label{fig:local_A*}
\end{figure}

A simple 2-D simulation demonstration is shown in Figure \ref{fig:local_A*}, where black blocks represent different obstacles, two red points are the starting point and the endpoint. As exhibited in the figure, when the robot encounters an obstacle that leads to zero new neighbors from the current sampler setting, sampler changes searching direction to move away from the direct path between the start state and goal state. Once front obstacles are absent, the planner return to searching the shortest path to the final goal from current state as expected. 

\subsection{Experiment for Comparison with Local Hybrid A*}
To compare the extension algorithm with the standard hybrid A* algorithm in the same benchmark, we controlled the searching steps of both cell A* and hybrid A* to compare the path difference between them under similar nodes visited. The experiment is implemented in MATLAB.

\begin{figure}[H]
    \centering
    \includegraphics[width=0.32\textwidth]{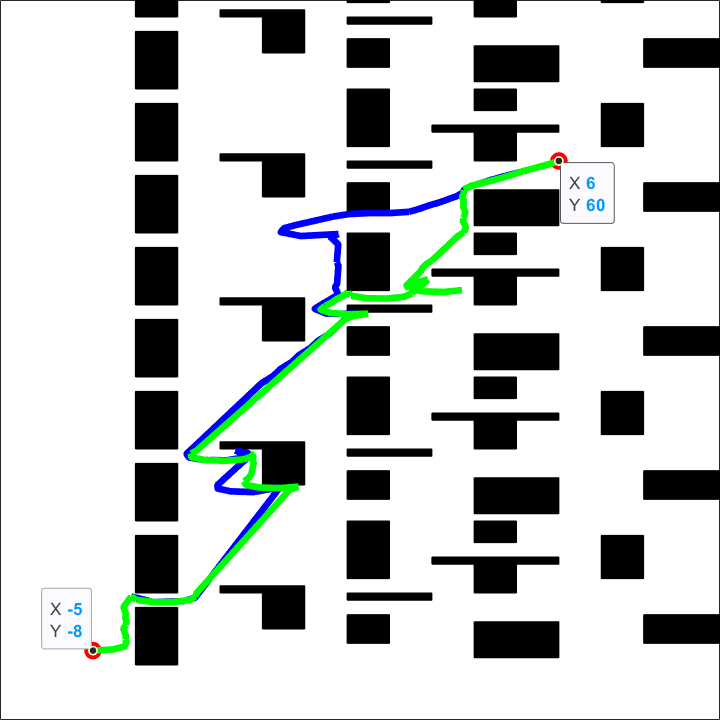}
    \includegraphics[width=0.32\textwidth]{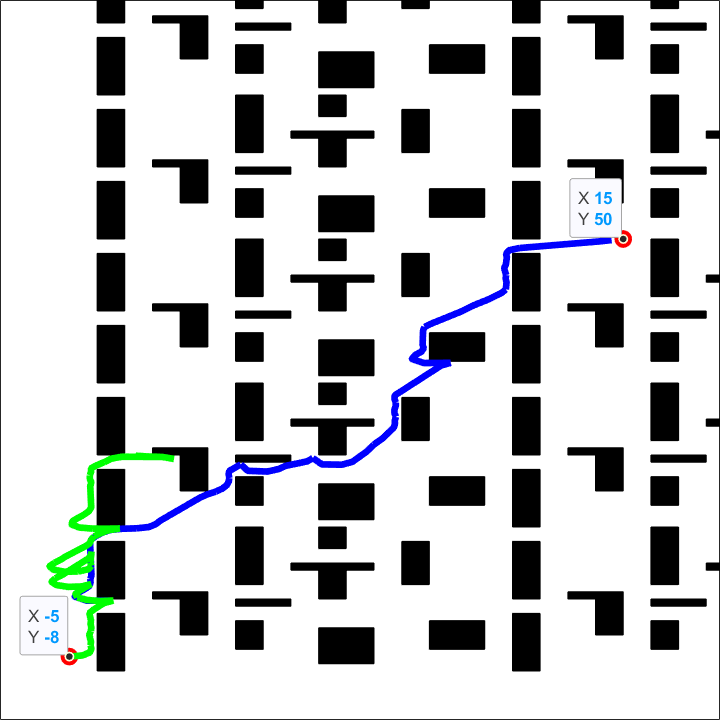}
    \includegraphics[width=0.32\textwidth]{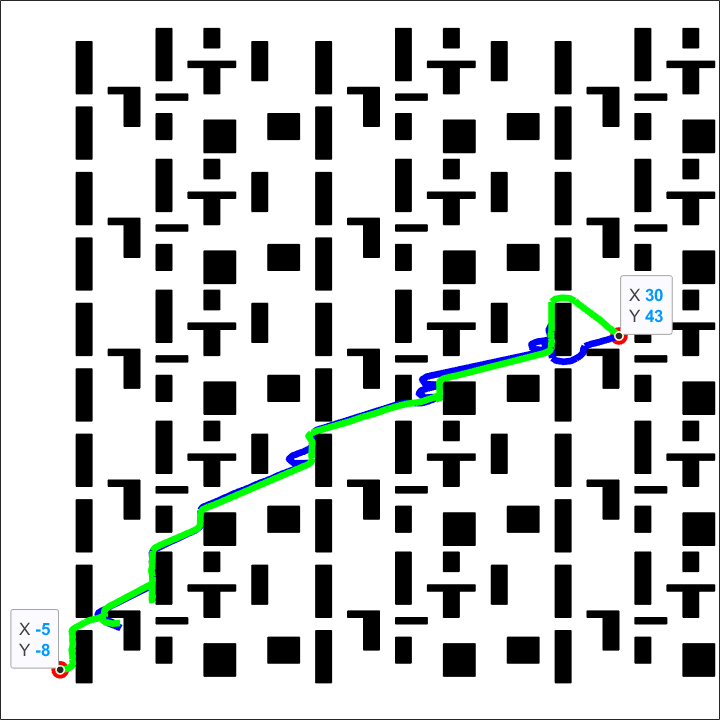}
    \caption{Cell A* (Blue) vs. Local Hybrid A* (Green) for four different destinations}
    \label{fig:comparison}
\end{figure}

As shown in Figure \ref{fig:comparison}, once hybrid A* was constrained within a small step of search, due to the limit of sampling depth, it failed to escape from obstacles in certain occasions like scenario with goal state $(15, 50)$. The robot started to move back to the trapped corner if the local nodes sampler fail to expand itself into a safe-to-pass area. Since cell A* adjusts sampling distance according to the current sampling situation, it can increase sampling distance to search deeper into the environment, hence increase the possibility of finding a path away from obstacles.

Table \ref{Adata} details how these two methods perform when they both head to the same goal position. "Failed" in Local Hybrid A* means robot ended up in a dead corner and failed to escape. From table, it can be inferred that cell A* needs less computation, and it is more stable compared to the local hybrid A* algorithm.

\begin{table}[h!t]
\centering
    \caption{Cell A* vs. Local hybrid A*}
\label{Adata}
\begin{tabular}{ |c|c|c|c|c| }
\hline
{Algorithm} & {Goal state} & {Time (sec)} & {Path Length (m)} & {Average Nodes}\\
\hline
{} & (6, 60) & 40.53 & 80.01 & 30.5 \\
{Cell A*} & (15, 50) & 46.33 & 91.93 & 27.79  \\
{} & (30, 43) & 43.67 & 87.08 & 27.96 \\ 
\hline
{} & (6, 60) & 47.43 & 86.5 & 31.1 \\
{Local Hybrid A*} & (15, 50) & Failed & Failed & Failed  \\
{} & (30, 43) & 53.293 & 91.5 & 29.56 \\ 
\hline
\end{tabular}
\end{table}

\subsection{Experiment for Comparison with D* lite}
\begin{figure}[h!t]
\centering
\includegraphics[width=0.32\textwidth]{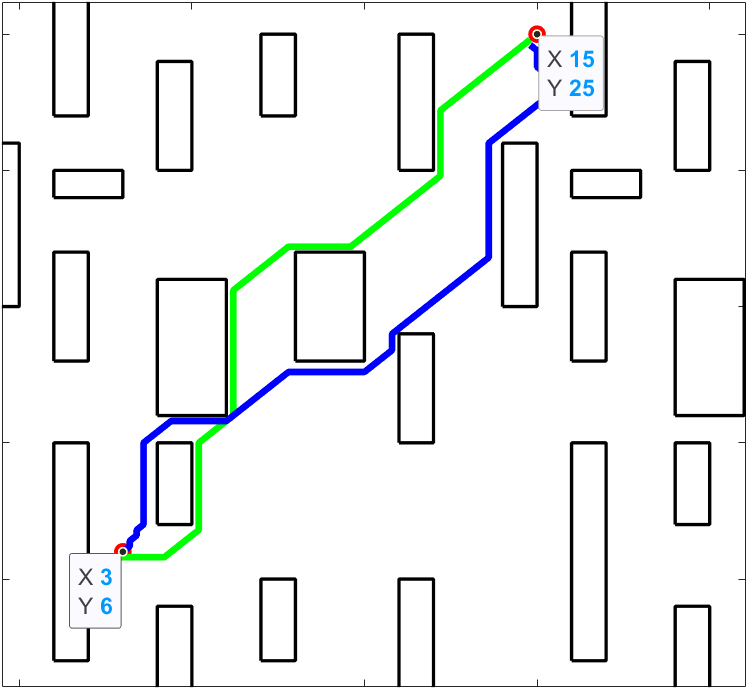}
\includegraphics[width=0.32\textwidth]{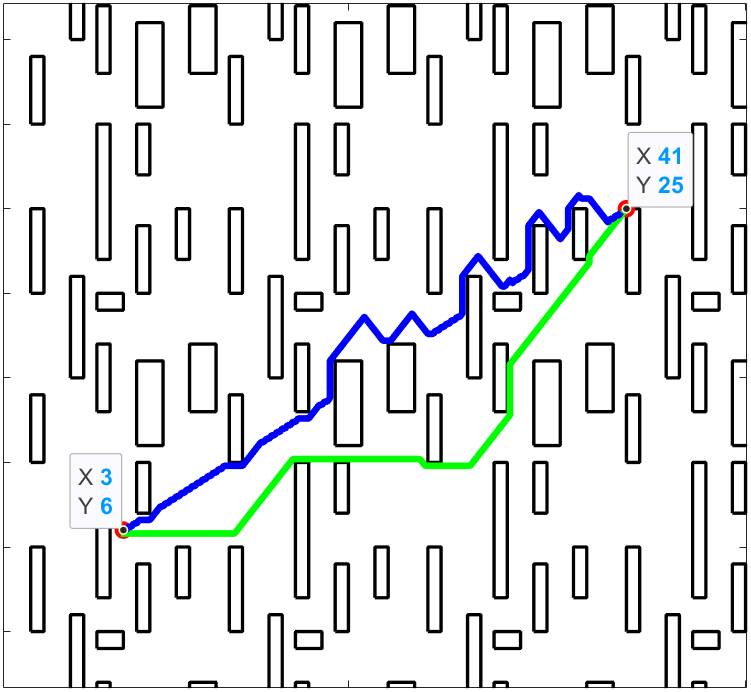}
\includegraphics[width=0.32\textwidth]{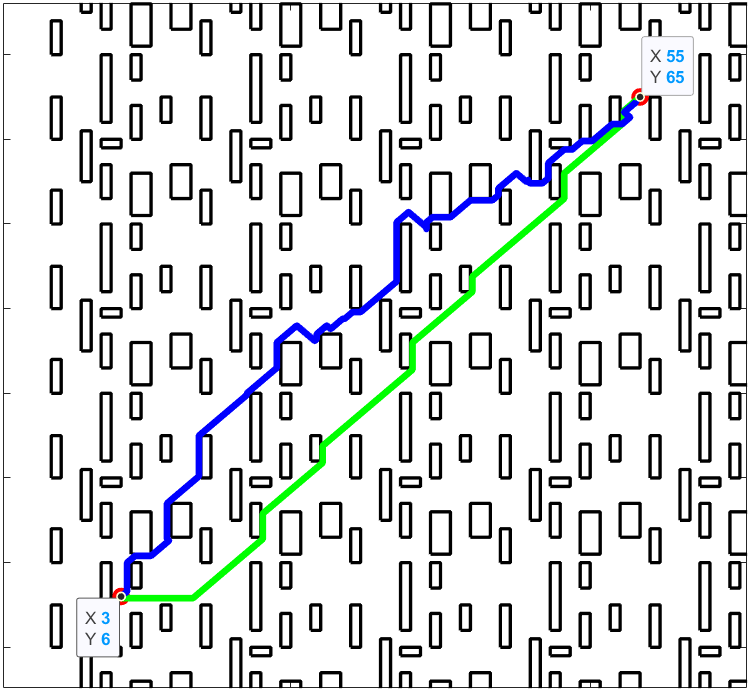}
\caption{Cell A* (Blue) vs. D* Lite (Green) in discrete map}
\label{fig:Cell_a_D_star}
\end{figure}

To further measure the navigation performance of Cell A*, a comparison between D* Lite and Cell A* in discrete state-space is implemented in MATLAB. In this test, the robot has 8 degrees of freedom of motion in a 2-D plane.
D* Lite is built based on dynamic programming; therefore, it can return the shortest path with any given 2-D map information at the cost of high time complexity. This comparison focuses on the overall traveling distance and planning time. The result is shown in Figure \ref{fig:Cell_a_D_star}.

As shown in the picture, Cell A* has a different path pattern compared to the path generated from D* Lite. To further analyze the difference, a detailed statistic is shown in Table \ref{dc_data}.

\begin{table}[h!t]
\centering
    \caption{Cell A* vs. D* lite}
\label{dc_data}
\begin{tabular}{ |c|c|c|c| }
\hline
{Algorithm} & {Goal state} & {Total Search Time (sec)} & {Path Length (number of nodes)}\\
\hline
{} & (15, 25) & 0.12 & 115 \\
{Cell A*} & (41, 25) & 0.13 & 230 \\
{} & (55, 65) & 0.25 & 403  \\
\hline
{} & (15, 25) & 1.45 & 110 \\ 
{D* lite} & (41, 25) & 6.12 & 206 \\
{} & (55, 65) & 2.07 & 331  \\
\hline
\end{tabular}
\end{table}

As shown in Table \ref{dc_data}, Cell's A* has much less computational time due to the strictly constrained search depth. However, the path nodes are not significantly larger than optimal solution from D* Lite, which indicates cell's A* can avoid some redundant searches from D* Lite.

\bigskip
\section{Algorithm Implement for 3D Case and Results}
This section exhibits the results of the algorithm implemented on a drone platform in a 3D environment. 
\subsection{Drones Execution Set-up}
\begin{figure}[h!t]
    \centering
    \includegraphics[width=0.6\textwidth]{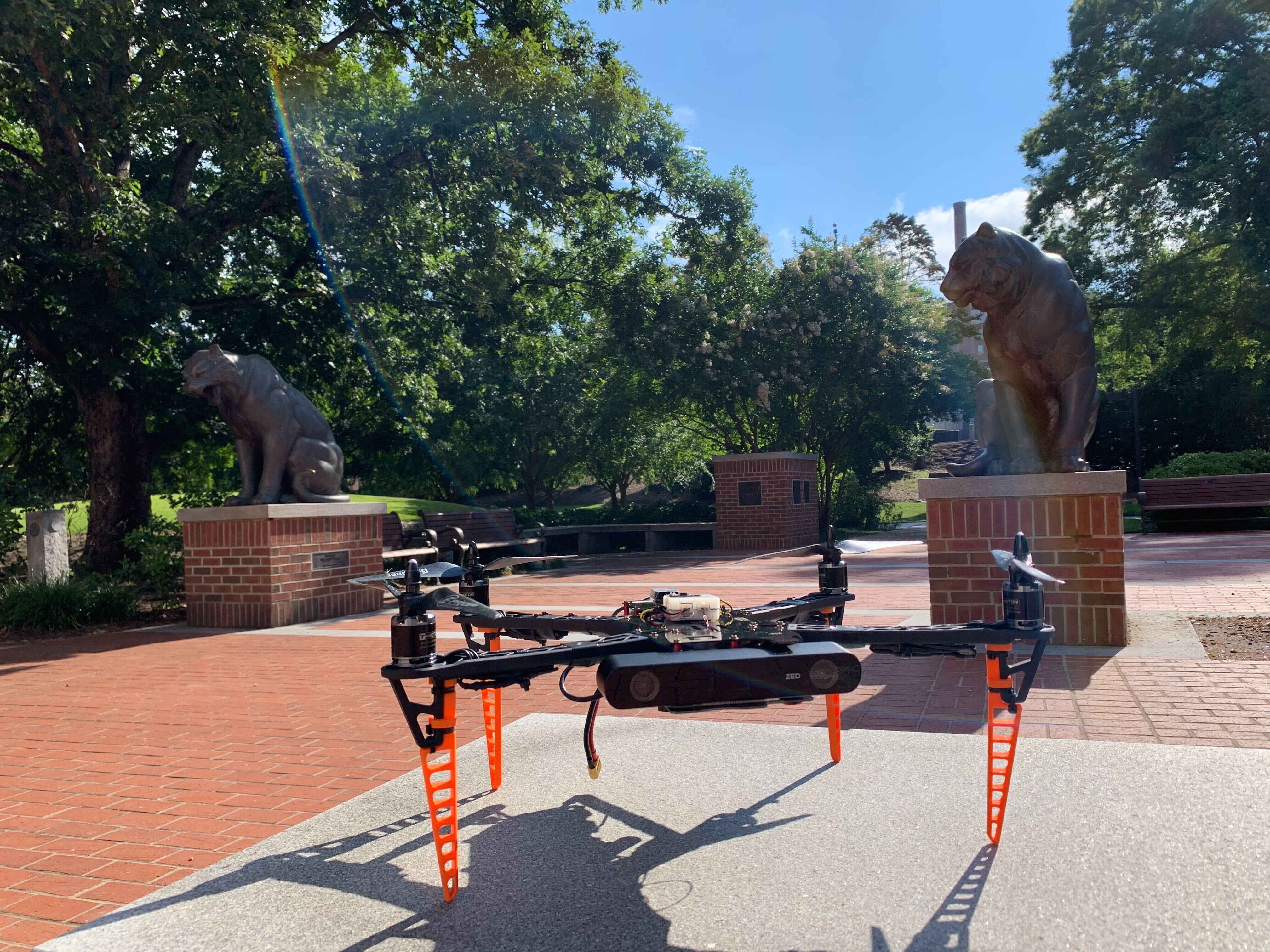}
    \caption{Modified TBS Discovery development platform.}
    \label{fig:physical}
\end{figure}

Before we deployed the algorithm in the real world, we tested it in a gazebo virtual environment built by GAAS group \cite{gaas}. The simulation is implemented in the robotics operating system (ROS) \cite{ros}. Localization is enabled by SLAM via a stereo camera sensor. For the real-world implementation, we modified a TBS Discovery drone frame to accommodate the on-board sensors, Pixracer PX4 autopilot, and NVIDIA TX2 single board computer. The TX2 build environment is configured using jetpack 4.2, which runs on ubuntu 18.04. The ROS build version is melodic. The communication between the high-level (e.g. Local pose updates from the navigation algorithm) and low-level control (flight controller) is done over MAVROS. The PX4 controller \cite{px4} takes care of the path following part in both simulation and real-world experiments in real-time. The above codes are all written with python language, and the physical drone we use is exhibited in Figure \ref{fig:physical}.
On-board sensors shown in Figure \ref{fig:physical} include a downward facing LiDAR for altitude measurement. The introduction of LiDAR provides accurate altitude reading by fusing the measurements with the on-board barometer readings. The on-board sensing module accurately tracks the drone's planar velocity using an optical flow sensor and is fused with IMU data to more accurately measure the speed of the drone. The forward-facing stereo camera (Zed 2 RGBD camera) is used to generate the local map data.    
\subsection{Experiment and Results}
\subsubsection{Experiment for Obstacles Avoidance}
\begin{figure}[h!t]
    \centering
    \includegraphics[width=0.6\textwidth]{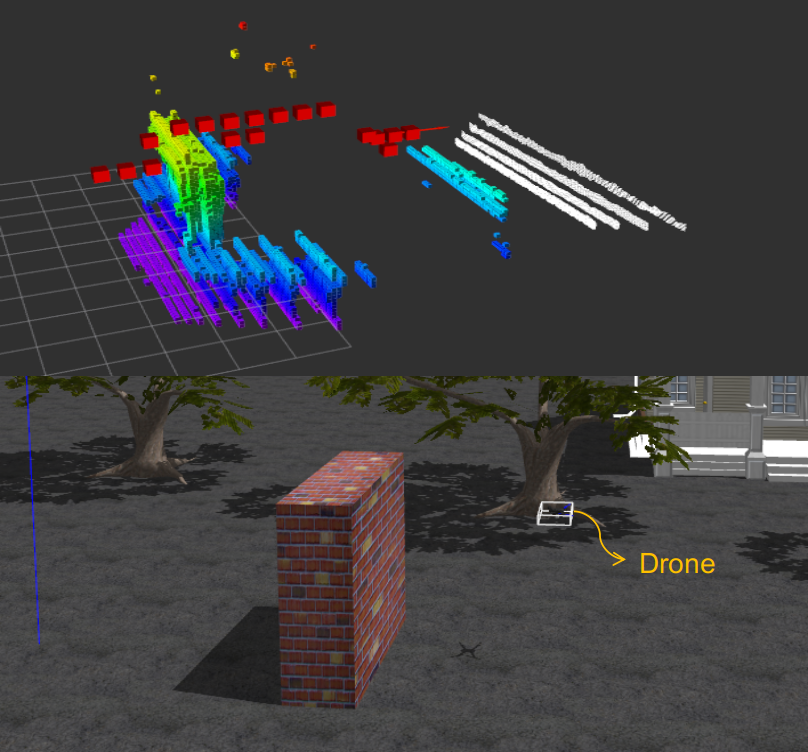}
    \caption{Obstacles avoidance test}
    \label{fig:test1}
\end{figure}

 In this experiment, the algorithm is tested on its ability to keep away from a static wall obstacle. The drone starts form position $(0, 0, 2.5)$ and heads to the goal position point $(7, 0, 2.5)$. The results is shown in Figure \ref{fig:test1}, where the red points are the recorded trajectory.
As shown in Figure \ref{fig:test1}, the drone correctly avoid the wall while it is within its defined obstacles avoidance range.

\subsubsection{Experiment for Long Distance Task}
\begin{figure}[h!t]
\centering
\includegraphics[width=0.6\textwidth]{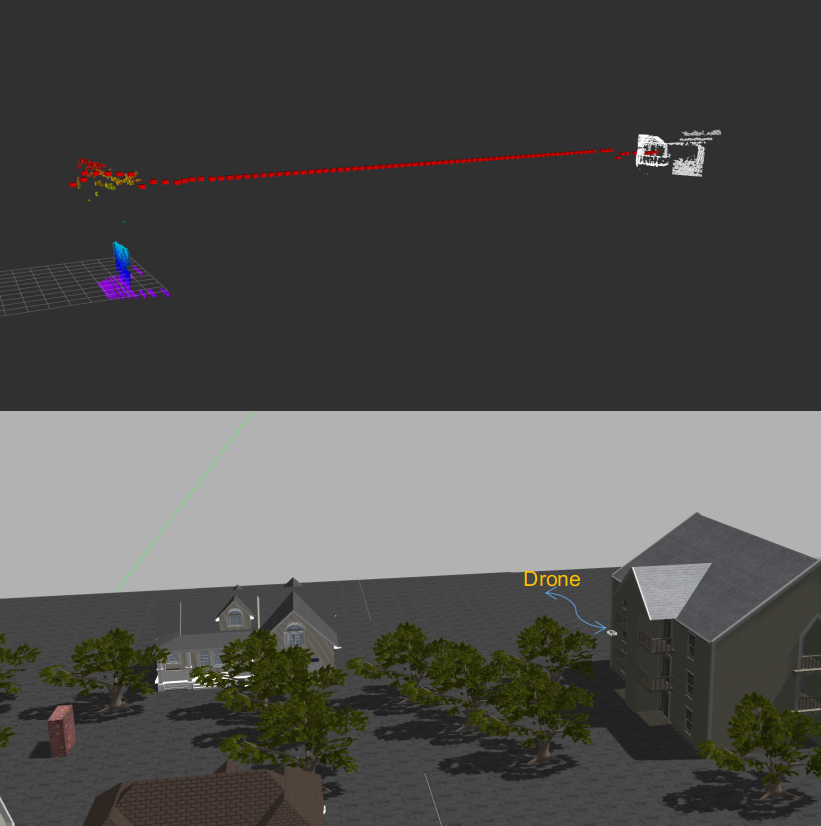}
\caption{Long distance task test}
\label{fig:test2}
\end{figure}

 In this experiment, the algorithm is tested on its ability for a long-distance path planning task. The drone takes off at position $(0, 0, 6)$ and the goal position is point $(45, -6, 8)$, the euclidean distance between these two points is $45.44m$. A planned route scheduled with the algorithm is exhibited in Figure \ref{fig:test2}. 
A clean path is shown in Figure \ref{fig:test2} that the drones avoid some trees at first and then flies straightly to the defined position, which is the balcony of the building. The top subplot in Figure \ref{fig:test2} shows the detected occupancy grids in a 3-D space. The 3D representation of the world is mostly sparse since we design Cell A* with efficiency as the top priority. We demonstrate that online trajectory planning can be achieved for an unknown environment.

\subsection{Experiment Results}
Table \ref{data} shows the results achieved by the two drone deployment experiments. It can be inferred that the drone simulation with the proposed algorithm can effectively handle unknown obstacles like a wall or foliage on trees during navigating in a semi-unknown environment.
\begin{table}[h!t]
\centering
    \caption{Experiment Data}
\label{data}
\begin{tabular}{ |c|c|c|c| }
\hline
Experiment & Task's Euclidean Distance (m) & Time (sec) & Exploit Nodes\\
\hline
{Obstacle avoidance} 
& 7 & 22.2319 & 13000  \\ 
\hline
{Long Distance} 
& 45.44 & 98.6619 & 91000  \\
\hline
\end{tabular}
\end{table}

\bigskip
\section{Conclusion and Future Work}
This paper presents an extension work of the Hybrid A* algorithm to enable UAVs run in a 3-D world or a car-like robot in a 2-D world without any prior information of the environment. By adjusting the nodal cost function, we allow the cost function to be a linear combination of the euclidean distance between the current location to the goal location and the distance between the current and the shortest euclidean path between the initial position and the goal position. In contrast, the sign of the second term is adaptive, depending on the obstacle in front of the planned path. The proposed Cell A* requires much less computing costs for the on-board computer. The algorithm requires at least one order of magnitude less sampling nodes than the Hybrid A* algorithm at each time step during online replanning. By taking advantage of the computational saving of the Cell A*, the drone in the simulation performs navigation with high efficiency not only in terms of obtaining a collision-free and smooth trajectory path but also saving much less time compared to the standard A* algorithm and its variants.  



\bibliographystyle{unsrt}
\bibliography{refs}

\end{document}